\documentclass[11pt,a4paper]{article}
\usepackage[hyperref]{acl2019}
\usepackage{times}
\usepackage{latexsym}
\usepackage{amsmath}
\usepackage{amssymb}
\usepackage{bbm}
\usepackage{booktabs}
\usepackage{multirow,multicol}
\usepackage{graphicx}
\usepackage{subfig}
\usepackage{color,soul}
\usepackage{tabularx}
\usepackage[export]{adjustbox}

\graphicspath{{figures}}

\usepackage{url}

\aclfinalcopy 

\title{\papertitle}

\author{Victor Zhong \\
  University of Washington \\
  \texttt{vzhong@cs.washington.edu} \\\And
  Luke Zettlemoyer \\
  University of Washington \\
  \texttt{lsz@cs.washington.edu} \\}
  
\date{}

\begin{document}
\newcommand{\tocite}[1]{{[\hl{CITE: #1]}}}
\newcommand{\todo}[1]{{[\hl{TODO: #1}]}}
\newcommand{\victor}[1]{{[\hl{VICTOR: #1}]}}
\newcommand{\luke}[1]{{[\hl{LUKE: #1]}}}

\newcommand{\papertitle}{{$\mathrm{E}^3$: Entailment-driven Extracting and Editing for Conversational Machine Reading}}
\newcommand{\modelname}{{Entailment-driven Extract and Edit network}}
\newcommand{\modelnameshort}{{$\mathrm{E}^3$}}
\newcommand{\sota}{{state-of-the-art}}

\newcommand{\sotadiff}{{3\%}}
\newcommand{\submissiondate}{{SUBMISSION DATE}}

\newcommand{\macroacc}{{73.3}}
\newcommand{\diffmacroacc}{{4.4}}
\newcommand{\microacc}{{67.6}}
\newcommand{\diffmicroacc}{{5.7}}
\newcommand{\diffmicroaccbert}{{4.0}}
\newcommand{\bleuone}{{54.1}}
\newcommand{\diffbleuone}{{-0.3}}
\newcommand{\bleufour}{{38.7}}
\newcommand{\diffbleufour}{{4.3}}
\newcommand{\diffbleufourbert}{{2.4}}
\newcommand{\combtest}{{28.4}}

\newcommand{\real}[1]{{\mathbb{R}^{{#1}}}}
\newcommand{\bigru}[1]{{\rm BiGRU} \left( {{#1}} \right)}
\newcommand{\coattn}[2]{{\rm Coattn} \left( {{#1}, {#2}} \right)}
\newcommand{\selfattn}[1]{{\rm Selfattn} \left( {{#1}} \right)}

\newcommand{\demb}{d_{\rm emb}}
\newcommand{\dhid}{d_{\rm hid}}

\newcommand\blfootnote[1]{%
  \begingroup
  \renewcommand\thefootnote{}\footnote{#1}%
  \addtocounter{footnote}{-1}%
  \endgroup
}
\maketitle
\begin{abstract}

Conversational machine reading systems help users answer high-level questions (e.g.~determine if they qualify for particular government benefits) when they do not know the exact rules by which the determination is made (e.g.~whether they need certain income levels or veteran status). The key challenge is that these rules are only provided in the form of a procedural text  (e.g.~guidelines from government website) which the system must read to figure out what to ask the user.
We present a new conversational machine reading model that jointly extracts a set of decision rules from the procedural text while reasoning about which are entailed by the conversational history and which still need to be edited to create questions for the user. On the recently introduced ShARC conversational machine reading dataset, our~\modelname~(\modelnameshort)~achieves a new~\sota, outperforming existing systems as well as a new BERT-based baseline.
In addition, by explicitly highlighting which information still needs to be gathered,~\modelnameshort~provides a more explainable alternative to prior work.
We release source code for our models and experiments at~\url{https://github.com/vzhong/e3}.
\end{abstract}

\section{Introduction}

\begin{figure}[!t]
    \centering
    \includegraphics[width=\linewidth]{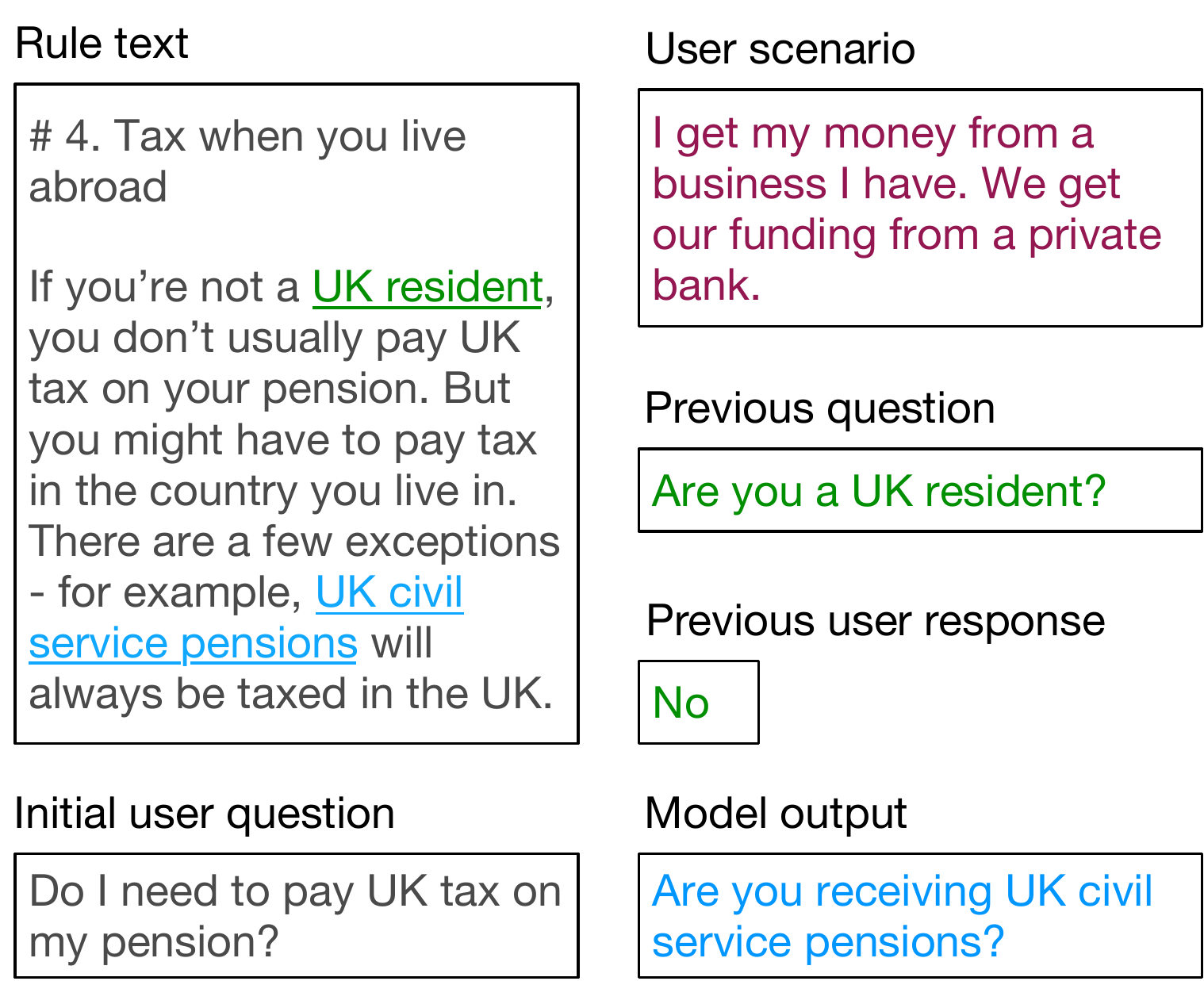}
    \caption{
A conversational machine reading example.
The model is given a rule text document, which contains a recipe of implicit rules (underlined) for answering the initial user question.
At the start of the conversation, the user presents a scenario describing their situation.
During each turn, the model can ask the user a follow-up question to inquire about missing information, or conclude the dialogue by answering \texttt{yes}, \texttt{no}, or \texttt{irrelevant}.
\texttt{irrelevant} means that the rule text cannot answer the question.
We show previous turns as well as the corresponding inquired rules in green.
The scenario is shown in red and in this case does not correspond to a rule.
The model inquiry for this turn and its corresponding rule are shown in blue.
    }
    \label{fig:example}
    \vspace{-0.2in}
\end{figure}

In conversational machine reading (CMR), a system must help users answer high-level questions by participating in an information gathering dialog. For example, in Figure~\ref{fig:example} the system asks a series of questions to help the user decide if they need to pay tax on their pension.
A key challenge in CMR is that the rules by which the decision is made are only provided in natural language (e.g.~the rule text in Figure~\ref{fig:example}). 
At every step of the conversation, the system must read the rules text and reason about what has already been said in to formulate the best next question.

We present a new model that jointly reasons about what rules are present in the text and which are already entailed by the conversational history to improve question generation. 
More specifically, we propose the~\modelname~(\modelnameshort).
\modelnameshort~learns to extract implicit rules in the document, identify which rules are entailed by the conversation history, and edit rules that are not entailed to create follow-up questions to the user.
During each turn,~\modelnameshort~parses the rule text to extract spans in the text that correspond to implicit rules (underlined in Figure~\ref{fig:example}).
Next, the model scores the degree to which each extracted rule is entailed by the initial user scenario (red in Figure~\ref{fig:example}) and by previous interactions with the user (green in Figure~\ref{fig:example}).
Finally, the model decides on a response by directly answering the question (\texttt{yes}/\texttt{no}), stating that the rule text does not contain sufficient information to answer the question (\texttt{irrelevant}), or asking a follow-up question about an extracted rule that is not entailed but needed to determine the answer (blue in Figure~\ref{fig:example}).
In the case of inquiry, the model edits an extracted rule into a follow-up question.
To our knowledge,~\modelnameshort~is the first extract-and-edit method for conversational dialogue, as well as the first method that jointly infers implicit rules in text, estimates entailment, inquires about missing information, and answers the question.

We compare~\modelnameshort~to the previous-best systems as well as a new, strong, BERT-based extractive question answering model (BERTQA) on the recently proposed ShARC CMR dataset~\citep{Saeidi2018interpretation}.
Our results show that~\modelnameshort~is more accurate in its decisions and generates more relevant inquiries.
In particular,~\modelnameshort~outperforms the previous-best model by~\diffmicroacc\% in micro-averaged decision accuracy and~\diffbleufour~in inquiry BLEU4.
Similarly,~\modelnameshort~outperforms the BERTQA baseline by~\diffmicroaccbert\% micro-averaged decision accuracy and~\diffbleufourbert~in inquiry BLEU4.
In addition to outperforming previous methods,~\modelnameshort~is explainable in the sense that one can visualize what rules the model extracted and how previous interactions and inquiries ground to the extracted rules.
We release source code for~\modelnameshort~and the BERTQA model at~\url{https://github.com/vzhong/e3}.

\section{Related Work}

\paragraph{Dialogue tasks.}
Recently, there has been growing interest in question answering (QA) in a dialogue setting~\citep{choi2018QuAC,reddy2019coqa}.
CMR~\citep{Saeidi2018interpretation} differs from dialogue QA in the domain covered (regulatory text vs Wikipedia).
A consequence of this is that CMR requires the interpretation of complex decision rules in order to answer high-level questions, whereas dialogue QA typically contains questions whose answers are directly extractable from the text.
In addition, CMR requires the formulation of free-form follow-up questions in order to identify whether the user satisfies decision rules, whereas dialogue QA does not.
There has also been significant work on task-oriented dialogue, where the system must inquire about missing information in order to help the user achieve a goal~\citep{dstc1,dstc2,mrkvsic2016neural,young2013POMDPDialogueReview}.
However, these tasks are typically constrained to a fixed ontology (e.g.~restaurant reservation), instead of a latent ontology specified via natural language documents.

\begin{figure*}[!t]
    \centering
    \includegraphics[width=\linewidth]{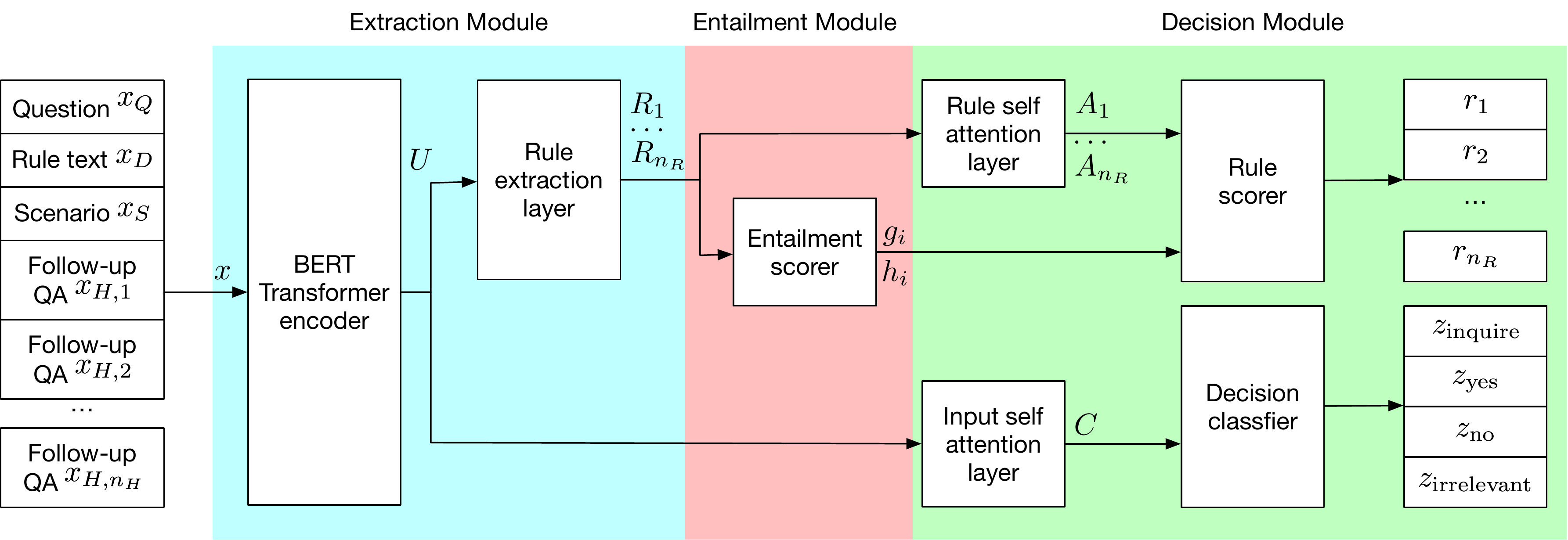}
    \caption{The~\modelname.}
    \label{fig:model}
    \vspace{-0.1in}
\end{figure*}

\paragraph{Dialogue systems.}
One traditional approach for designing dialogue systems divides the task into language understanding/state-tracking~\citep{mrkvsic2016neural,zhong2018global}, reasoning/policy learning~\citep{Su2016ActiveReward}, and response generation~\citep{Wen2015SemanticConditionedGen}.
The models for each of these subtasks are then combined to form a full dialogue system~\citep{young2013POMDPDialogueReview,wen2017NetworkBasedEndToEndDialogueSystem}.
The previous best system for ShARC~\citep{Saeidi2018interpretation} similarly breaks the CMR task into subtasks and combines hand-designed sub-models for decision classification, entailment, and follow-up generation.
In contrast, the core reasoning (e.g. non-editor) components of~\modelnameshort~are jointly trained, and does not require complex hand-designed features.

\paragraph{Extracting latent rules from text.}
There is a long history of work on extracting knowledge automatically from text~\citep{Moulin1992Automated}.
Relation extraction typically assumes that there is a fixed ontology onto which extracted knowledge falls~\citep{Mintz2009DistantSupRE,riedel2013REMatFac}.
Other works forgo the ontology by using, for example, natural language~\citep{angeli2014Naturalli,angeli2015openie}.
These extractions from text are subsequently used for inference over a knowledge base~\citep{bordes2013TransE,dettmers2018conve,lin2018multihopKG} and rationalizing model predictions~\citep{lei2016rationalizing}.
Our work is more similar with the latter type in which knowledge extracted are not confined to a fixed ontology and instead differ on a document basis.
In addition, the rules extracted by our model are used for inference over natural language documents.
Finally, these rules provide rationalization for the model's decision making, in the sense that the user can visualize what rules the model extracted and which rules are entailed by previous turns.


\section{\modelname}

In conversational machine reading, a system reads a document that contains a set of implicit decision rules.
The user presents a scenario describing their situation, and asks the system an underspecified question.
In order to answer the user's question, the system must ask the user a series of follow-up questions to determine whether the user satisfies the set of decision rules.

The key challenges in CMR are to identify implicit rules present in the document, understand which rules are necessary to answer the question, and inquire about necessary rules that are not entailed by the conversation history by asking follow-up questions.
The three core modules of~\modelnameshort, the extraction, entailment, and decision modules, combine to address these challenges.
Figure~\ref{fig:model} illustrates the components of~\modelnameshort.

For ease of exposition, we describe~\modelnameshort~for a single turn in the conversation.
To make the references concrete in the following sections, we use as an example the inputs and outputs from Figure~\ref{fig:example}.
This example describes a turn in a conversation in which the system helps the user determine whether they need to pay UK taxes on their pension.

\subsection{Extraction module}
\label{sec:extraction}
The extraction module extracts spans from the document that correspond to latent rules.
Let $x_D$, $x_Q$, $x_S$, $x_{H,i}$ denote words in the rule text, question, scenario, and the inquiry and user response during the $i$th previous turn of the dialogue after $N$ turns have passed.
We concatenate these inputs into a single sequence $x = [x_Q; x_D; x_S; x_{H,1}; \cdots x_{H,N}]$ joined by sentinel tokens that mark the boundaries of each input.
To encode the input for the extraction module, we use BERT, a transformer-based model~\citep{vaswani2017Attention} that achieves consistent gains on a variety of NLP tasks~\citep{Devlin2018BERT}.
We encode $x$ using the BERT encoder, which first converts words into word piece tokens~\citep{Wu2016GooglesNM}, then embeds these tokens along with their positional embeddings and segmentation embeddings.
These embeddings are subsequently encoded via a transformer network, which allows for inter-token attention at each layer.
Let $n_x$ be the number of tokens in the concatenated input $x$ and $d_U$ be the output dimension of the BERT encoder.
For brevity, we denote the output of the BERT encoder as $U = \mathrm{BERT}(x) \in \real{n_x \times d_U}$ and refer readers to~\citet{Devlin2018BERT}~for detailed architecture.

In order to extract the implicit decision rules from the document, we compute a start score $\alpha_i$ and an end score $\beta_i$ for each $i$th token as
\begin{eqnarray}
\label{eq:extraction_prob}
\alpha_i &=& \sigma \left( W_\alpha U_i + b_\alpha \right) \in \real{}
\\
\beta_i &=& \sigma \left( W_\beta U_i + b_\beta \right) \in \real{}
\end{eqnarray}
where $W_\alpha, W_\beta \in \real{d_U}$,
$b_\alpha, b_\beta \in \real{}$,
and $\sigma$ is the sigmoid function.

For each position $s_i$ where $\alpha_i$ is larger than some threshold $\tau$, we find the closest proceeding position $e_i \ge s_i$ where $\beta_{e_i} > \tau$.
Each pair $(s_i, e_i)$ then forms an extracted span corresponding to a rule $R_i$ expressed in the rule text.
In the example in Figure~\ref{fig:example}, the correct extracted spans are ``UK resident'' and ``UK civil service pensions''.

For the $i$th rule, we use self-attention to build a representation $\overline{A}_i$ over the span $(s_i, e_i)$.
\begin{eqnarray}
\overline{\gamma}_k &=& W_\gamma U_k + b_\gamma \in \real{}, s_i \le k \le e_i\\
\gamma_k &=& \mathrm{softmax}\left(\overline{\gamma}\right)_k \in \real{}, s_i \le k \le e_i\\
\overline{A}_i &=& \sum_{k=s_i}^{e_i} \gamma_k U_k \in \real{d_U}
\end{eqnarray}
where $W_\gamma \in \real{d_U}$ and $b_\gamma \in \real{}$.
Here, $\overline{\gamma}_k, \gamma_k$ are respectively the unnormalized and normalized scores for the self-attention layer.

Let $n_R$ denote the number spans in the rule text, each of which corresponds to a ground truth rule.
The rule extraction loss is computed as the sum of the binary cross entropy losses for each rule $R_i$.
\begin{eqnarray}
L_\mathrm{re} = \sum_i^{n_R} L_{\mathrm{start}, i} + L_{\mathrm{end}, i}
\end{eqnarray}

Let $n_D$ denote the number of tokens in the rule text, $s_i$, $e_i$ the ground truth start and end positions for the $i$th rule, and $\mathbbm{1}_f$ the indicator function that returns 1 if and only if the condition $f$ holds.
Recall from Eq~\eqref{eq:extraction_prob} that $\alpha_j$ and $\beta_j$ denote the probabilities that token $j$ is the start and end of a rule. 
The start and end binary cross entropy losses for the $i$th rule are computed as
\begin{eqnarray*}
L_{\mathrm{start}, i} = - \sum_j^{n_D} 
\mathbbm{1}_{j=s_i} \log\left(\alpha_j\right) + \mathbbm{1}_{j\neq s_i} \log\left( 1 - \alpha_j \right)\\
L_{\mathrm{end}, i} = - \sum_j^{n_D} 
\mathbbm{1}_{j=e_i} \log\left(\beta_j\right) + \mathbbm{1}_{j\neq e_i} \log\left( 1 - \beta_j \right)
\end{eqnarray*}

\subsection{Entailment module}

Given the extracted rules $R = \{R_1, \cdots R_{n_R}\}$, the entailment module estimates whether each rule is entailed by the conversation history, so that the model can subsequently inquire about rules that are not entailed.
For the example in Figure~\ref{fig:example}, the rule ``UK resident'' is entailed by the previous inquiry ``Are you a UK resident''.
In contrast, the rule ``UK civil service pensions'' is not entailed by either the scenario or the conversation history, so the model needs to inquire about it.
In this particular case the scenario does not entail any rule.

For each extracted rule, we compute a score that indicates the extent to which this particular rule has already been discussed in the initial scenario $S$ and in previous turns $Q$.
In particular, let $N(R_i, S)$ denote the number of tokens shared by $R_i$ and $S$, $N(R_i)$ the number of tokens in $R_i$, and $N(S)$ the number of tokens in $S$.
We compute the scenario entailment score $g_i$ as
\begin{eqnarray}
\mathrm{pr}(R_i, S) &=& \frac{N(R_i, S)}{N(R_i)}\\
\mathrm{re}(R_i, S) &=& \frac{N(R_i, S)}{N(S)}\\
g_i = \mathrm{f1}(R_i, S)
&=& \frac{
2
\mathrm{pr}(R_i, S)
\mathrm{re}(R_i, S)
}{
\mathrm{pr}(R_i, S) +
\mathrm{re}(R_i, S)
}
\end{eqnarray}
where $\mathrm{pr}$, $\mathrm{re}$, and $\mathrm{f1}$ respectively denote the precision, recall, and F1 scores.
We compute a similar score to represent the extent to which the rule $R_i$ has been discussed in previous inquiries.
Let $Q_k$ denote tokens in the $k$th previous inquiry.
We compute the history entailment score $h_i$ between the extracted rule $R_i$ and all $n_H$ previous inquiries in the conversation history as
\begin{eqnarray}
h_i = \max_{k = 1, \cdots n_H} \mathrm{f1}(R_i, Q_k)
\end{eqnarray}
The final representation of the $i$th rule, $A_i$, is then the concatenation of the span self-attention and the entailment scores.
\begin{eqnarray}
A_i &=& [\overline{A}_i; g_i; h_i] \in \real{d_U + 2}
\end{eqnarray}
where $[x; y]$ denotes the concatenation of $x$ and $y$.
We also experiment with embedding and encoding similarity based approaches to compute entailment, but find that this F1 approach performs the best. 
Because the encoder utilizes cross attention between different components of the input, the representations $U$ and $\overline{A}_i$ are able to capture notions of entailment.
However, we find that explicitly scoring entailment via the entailment module further discourages the model from making redundant inquiries.

\subsection{Decision module}

Given the extracted rules $R$ and the entailment-enriched representations for each rule $A_i$, the decision module decides on a response to the user.
These include answering \texttt{yes}/\texttt{no} to the user's original question, determining that the rule text is \texttt{irrelevant} to the question, or inquiring about a rule that is not entailed but required to answer the question.
For the example in Figure~\ref{fig:example}, the rule ``UK civil service pensions'' is not entailed, hence the correct decision is to ask a follow-up question about whether the user receives this pension.

We start by computing a summary $C$ of the input using self-attention
\begin{eqnarray}
\overline{\phi}_k &=& W_\phi U_k + b_\phi \in \real{}\\
\phi_k &=& \mathrm{softmax}\left( \overline{\phi} \right)_k \in \real{}\\
C &=& \sum_{k=s_i}^{e_i} \phi_k U_k \in \real{d_U}
\end{eqnarray}
where $W_\phi \in \real{d_U}$, $b_\phi \in \real{}$, and $\overline{\phi}$, $\phi$ are respectively the unnormalized and normalized self-attention weights.
Next, we score the choices \texttt{yes}, \texttt{no}, \texttt{irrelevant}, and \texttt{inquire}.
\begin{eqnarray}
z &=& W_z C + b_z \in \real{4}
\end{eqnarray}
where $z$ is a vector containing a class score for each of the \texttt{yes}, \texttt{no}, \texttt{irrelevant}, and \texttt{inquire} decisions.

For inquiries, we compute an inquiry score $r_i$ for each extracted rule $R_i$.
\begin{eqnarray}
r_i = W_z A_i + b_z \in \real{}
\end{eqnarray}
where $W_z \in \real{d_U + 2}$ and $b_z \in \real{}$.
Let $k$ indicate the correct decision, and $i$ indicate the correct inquiry, if the model is supposed to make an inquiry.
The decision loss is
\begin{eqnarray}
\label{eq:decision}
L_\mathrm{dec} &=& - \log \mathrm{softmax}(z)_k\\
&& -\mathbbm{1}_{k = \mathrm{inquire}} \log \mathrm{softmax}(r)_i \nonumber
\end{eqnarray}

During inference, the model first determines the decision $d = \mathrm{argmax}_k z_k$.
If the decision $d$ is \texttt{inquire}, the model asks a follow-up question about the $i$th rule such that $i = \mathrm{argmax}_j r_j$.
Otherwise, the model concludes the dialogue with $d$.

\paragraph{Rephrasing rule into question via editor.}

\begin{figure}[!t]
    \centering
    \includegraphics[width=\linewidth]{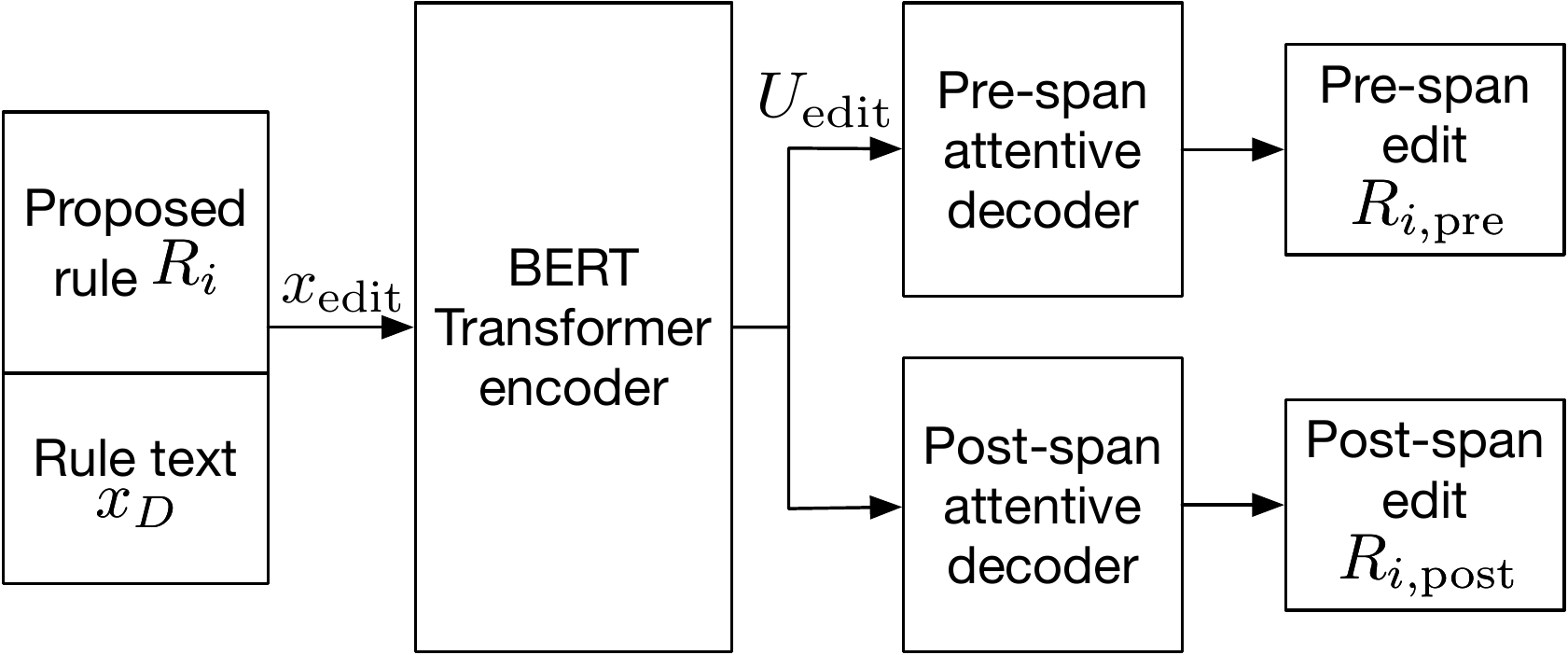}
    \caption{The editor of~\modelnameshort.}
    \label{fig:editor}
    \vspace{-0.2in}
\end{figure}

In the event that the model chooses to make an inquiry about an extracted rule $R_i$,
$R_i$ is given to an subsequent editor to rephrase into a follow-up question.
For the example in~\ref{fig:example}, the editor edits the span ``UK civil service pensions'' into the follow-up question ``Are you receiving UK civil service pensions?''
Figure~\ref{fig:editor} illustrates the editor.

The editor takes as input $x_\mathrm{edit} = [R_i; x_D]$, the concatenation of the extracted rule to rephrase $R_i$ and the rule text $x_D$.
As before, we encode using a BERT encoder to obtain $U_\mathrm{edit} = \mathrm{BERT}(x_\mathrm{edit})$.
The encoder is followed by two decoders that respective generate the pre-span edit $R_{i, \mathrm{pre}}$ and post-span edit $R_{i, \mathrm{post}}$.
For the example in Figure~\ref{eq:extraction_prob}, given the span ``UK civil service pensions'', the pre-span and post span edits that form the question ``Are you receiving UK civil service pensions?'' are respectively ``Are you receiving'' and ``?''

To perform each edit, we employ an attentive decoder~\citep{bahdanau2014neural} with Long Short-Term Memory (LSTM)~\citep{Hochreiter1997LongSM}.
Let $h_t$ denote the decoder state at time $t$.
We compute attention $a_t$ over the input.
\begin{eqnarray}
\label{eq:edit}
\overline{\zeta}_k &=& U_\mathrm{edit} h_{t-1} \in \real{}\\
\zeta_k &=& \mathrm{softmax}(\overline{\zeta})_k \in \real{}\\
a_t &=& \sum_k \zeta_k U_{\mathrm{edit},k} \in \real{d_U}
\end{eqnarray}

Let $V \in \real{n_V \times d_V}$ denote the embedding matrix corresponding to $n_V$ tokens in the vocabulary.
To generate the $t$th token $w_t$, we use weight tying between the output layer and the embedding matrix~\citep{press2017WeightTying}.
\begin{eqnarray}
v_t &=& \mathrm{embed}(V, w_{t-1})\\
h_t &=& \mathrm{LSTM}\left( [v_t; a_t], h_{t-1} \right) \in \real{d_U}\\
o_t &=& W_o [h_t; a_t] + b_o \in \real{d_V}\\
p(w_t) &=& \mathrm{softmax}(V o_t) \in \real{n_V}\\
w_t &=& \mathrm{argmax}_k p(w_t)_k
\end{eqnarray}

\begin{table*}[t]
\centering
\begin{tabularx}{\textwidth}{@{}XXXXXX@{}}
\toprule
Model & Micro Acc. & Macro Acc. & BLEU1 & BLEU4 & Comb. \\
\midrule
Seq2Seq & 44.8 & 42.8 & 34.0 & 7.8 & 3.3 \\
Pipeline & 61.9 & 68.9 & \textbf{54.4} & 34.4 & 23.7 \\
BERTQA & 63.6 & 70.8 & 46.2 & 36.3 & 25.7 \\
\modelnameshort~ (ours) & \textbf{\microacc} & \textbf{\macroacc} & \bleuone & \textbf{\bleufour} & \textbf{\combtest} \\
\bottomrule
\end{tabularx}
\caption{
Model performance on the blind, held-out test set of ShARC.
The evaluation metrics are micro and macro-averaged accuracy in classifying bewteen the decisions \texttt{yes}, \texttt{no}, \texttt{irrelevant}, and \texttt{inquire}.
In the event of an inquiry, the generated follow-up question is further evaluated using the BLEU score.
In addition to official evaluation metrics, we also show a combined metric (``Comb.''), which is the product between the macro-averaged accuracy and the BLEU4 score.
}
\label{tab:result}
\vspace{-0.1in}
\end{table*}

We use a separate attentive decoder to generate the pre-span edit $R_{i, \mathrm{pre}}$ and the post-span edit $R_{i, \mathrm{post}}$.
The decoders share the embedding matrix and BERT encoder but do not share other parameters.
The output of the editor is the concatenation of tokens $[R_{i, \mathrm{pre}}; R_i; R_{i, \mathrm{post}}]$.

The editing loss consists of the sequential cross entropy losses from generating the pre-span edit and the post-span edit.
Let $n_\mathrm{pre}$ denote the number of tokens and $\hat{w}_{t, \mathrm{pre}}$ the $t$th tokens in the ground truth pre-span edit.
The pre-span loss is
\begin{eqnarray}
\label{eq:prespan_loss}
L_\mathrm{pre} = - \sum_t^{n_\mathrm{pre}} \log p(\hat{w}_{t, \mathrm{pre}})
\end{eqnarray}

The editing loss is then the sum of the pre-span and post-span losses, the latter of which is obtained in a manner similar to Eq~\eqref{eq:prespan_loss}.
\begin{eqnarray}
L_\mathrm{edit} = L_\mathrm{pre} + L_\mathrm{post}
\end{eqnarray}


\section{Experiment}

We train and evaluate the~\modelname~on the ShARC CMR dataset.
In particular, we compare our method to three other models.
Two of these models are proposed by~\citet{Saeidi2018interpretation}.
They are an attentive sequence-to-sequence model that attends to the concatenated input and generates the response token-by-token (Seq2Seq), and a strong hand-engineered pipeline model with sub-models for entailment, classification, and generation (Pipeline).
For the latter,~\citet{Saeidi2018interpretation} show that these sub-models outperform neural models such as the entailment model by~\citet{parikh2016dam}, and that the combined pipeline outperforms the attentive sequence-to-sequence model.
In addition, we propose an extractive QA baseline based on BERT (BERTQA).
Similar models achieved~\sota~on a variety of QA tasks~\citep{Rajpurkar2016SQuAD10,reddy2019coqa}.
We refer readers to Section~\ref{sec:bertqa} of the appendices for implementation details BERTQA.

\subsection{Experimental setup}
We tokenize using revtok\footnote{\url{https://github.com/jekbradbury/revtok}} and part-of-speech tag (for the editor) using Stanford CoreNLP~\cite{Manning2014TheSC}.
We fine-tune the smaller, uncased pretrained BERT model by~\citet{Devlin2018BERT} (e.g.~\texttt{bert-base-uncased}).\footnote{We use the BERT implementation from \url{https://github.com/huggingface/pytorch-pretrained-BERT}}
We optimize using ADAM~\citep{Kingma2014AdamAM} with an initial learning rate of 5e-5 and a warm-up rate of 0.1.
We regularize using Dropout~\citep{Srivastava14dropout} after the BERT encoder with a rate of 0.4.

To supervise rule extraction, we reconstruct full dialogue trees from the ShARC training set and extract all follow-up questions as well as bullet points from each rule text and its corresponding dialogue tree.
We then match these extracted clauses to spans in the rule text, and consider these noisy matched spans as supervision for rule extraction.
During inference, we use heuristic bullet point extraction\footnote{We extract spans from the text that starts with the ``*'' character and ends with another ``*'' character or a new line.}
in conjunction with spans extracted by the rule extraction module.
This results in minor performance improvements (~$\sim1$\% micro/macro acc.) over only relying on the rule extraction module.
In cases where one rule fully covers another, we discard the covered shorter rule.
Section~\ref{sec:matching} details how clause matching is used to obtain noisy supervision for rule extraction.

We train the editor separately, as jointly training with a shared encoder worsens performance.
The editor is trained by optimizing $L_\mathrm{edit}$ while the rest of the model is trained by optimizing $L_\mathrm{dec} + \lambda L_\mathrm{re}$.
We use a rule extraction threshold of $\tau=0.5$ and a rule extraction loss weight of $\lambda=400$.
We perform early stopping using the product of the macro-averaged accuracy and the BLEU4 score.

\begin{figure*}[!t]
     \subfloat[\label{fig:pred1}]{%
       \includegraphics[width=0.49\textwidth,valign=t]{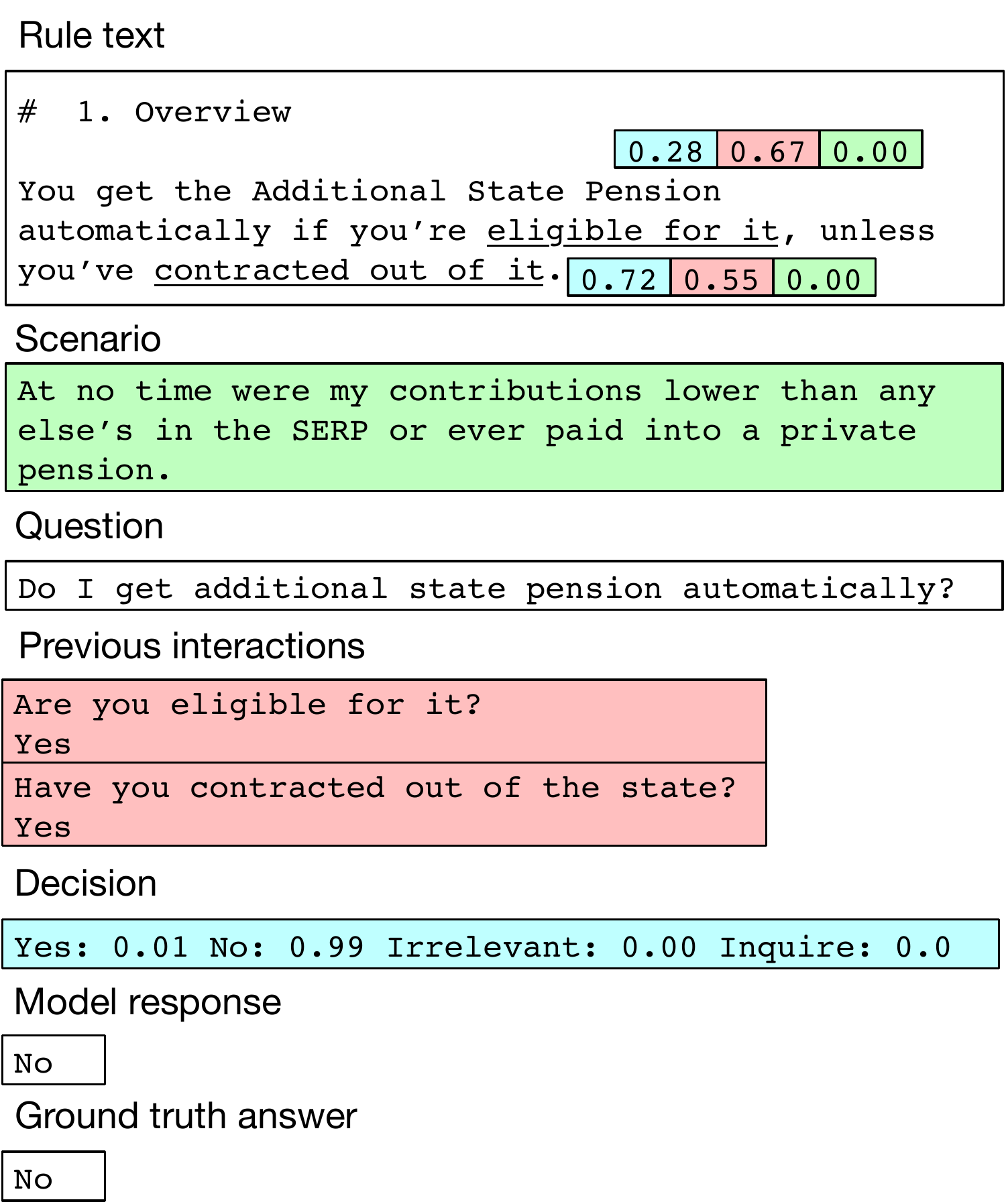}
     }
     \hfill
     \subfloat[\label{fig:pred3}]{%
       \includegraphics[width=0.49\textwidth,valign=t]{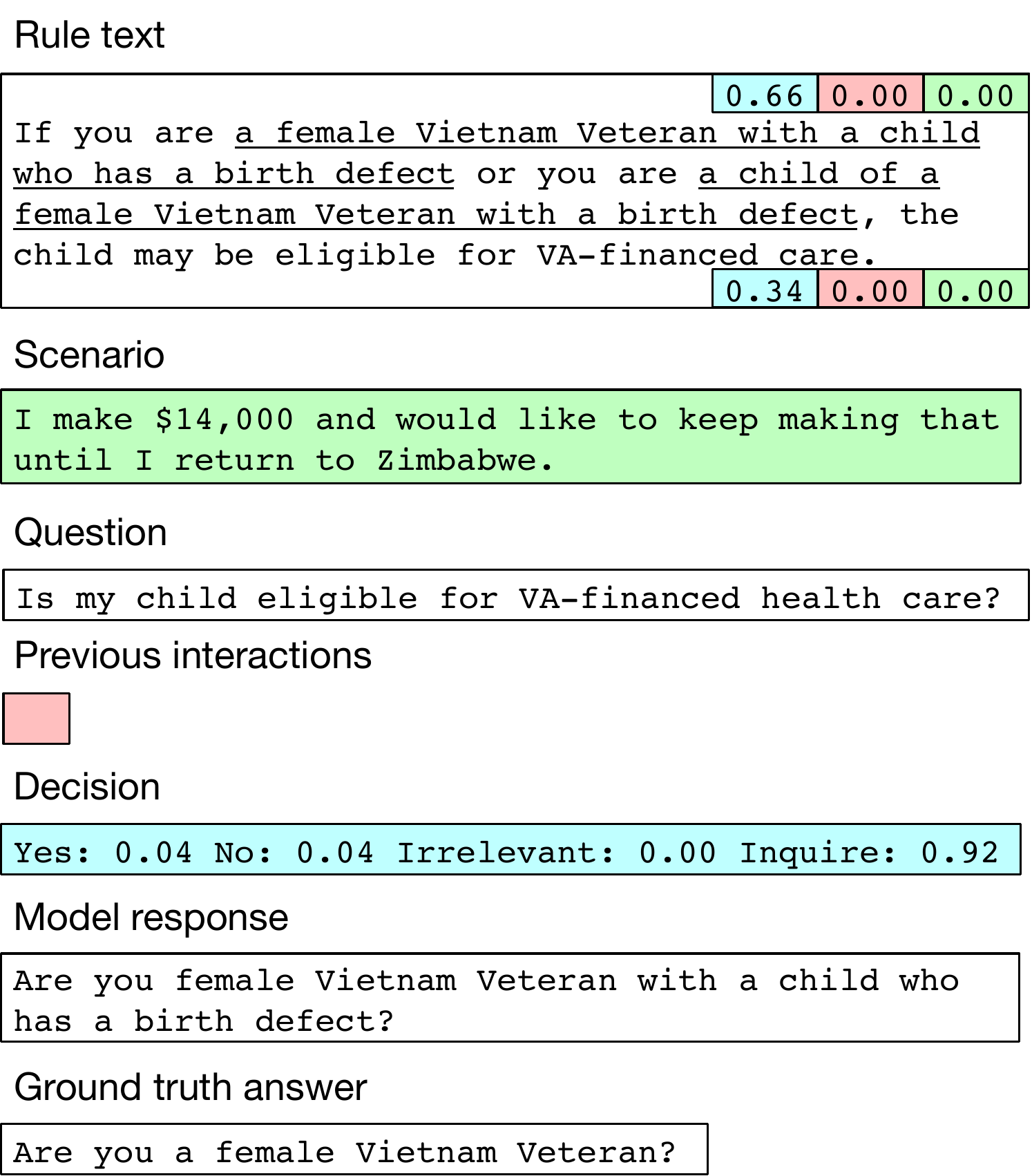}
     }
     \caption{
Predictions by~\modelnameshort.
Extracted spans are underlined in the text.
The three scores are the inquiry score $r_i$ (blue), history entailment score $h_i$ (red), and scenario entailment score $g_i$ (green) of the nearest extracted span.
    }
    \label{fig:preds}
\end{figure*}

For the editor, we use fixed, pretrained embeddings from GloVe~\citep{Pennington2014GloveGV}, and use dropout after input attention with a rate of 0.4.
Before editing retrieved rules, we remove prefix and suffix adpositions, auxiliary verbs, conjunctions, determiners, or punctuation.
We find that doing so allows the editor to convert some extracted rules (e.g.~or sustain damage) into sensible questions (e.g.~did you sustain damage?).

\subsection{Results}

Our performance on the development and the blind, held-out test set of ShARC is shown in Table~\ref{tab:result}.
Compared to previous results,~\modelnameshort~achieves a new~\sota, obtaining best performance on micro and macro-averaged decision classification accuracy and BLEU4 scores while maintaining similar BLEU1 scores.
These results show that~\modelnameshort~both answers the user's original question more accurately, and generates more coherent and relevant follow-up questions.
In addition, Figure~\ref{fig:preds}~shows that because~\modelnameshort~explicitly extracts implicit rules from the document, the model's predictions are explainable in the sense that the user can verify the correctness of the extracted rules and observe how the scenario and previous interactions ground to the extracted rules.

\begin{table*}[!t]
\centering
\begin{tabularx}{\textwidth}{@{}lXXXXX@{}}
\toprule
Model & Micro Acc. & Macro Acc. & BLEU1 & BLEU4 & Comb. \\
\midrule
\modelnameshort & 68.0 & 73.4 & 66.9 & 53.7 & 39.4 \\
-edit & 68.0 & 73.4 & 53.1 & 46.2 & 33.9 \\
-edit, entail & 68.0 & 73.1 & 50.2 & 40.3 & 29.5 \\
-edit, entail, extract (BERTQA) & 63.4 & 70.6 & 47.4 & 37.4 & 23.7 \\
\bottomrule
\end{tabularx}
\caption{
Ablation study of~\modelnameshort~on the development set of ShARC.
The ablated variants of~\modelnameshort~include versions:
without the editor;
without the editor and entailment module;
without the editor, entailment module, and extraction module, which reduces to the BERT for question answering model by~\citet{Devlin2018BERT}.
}
\label{tab:ablation}
\vspace{-0.1in}
\end{table*}

\subsection{Ablation study}

Table~\ref{tab:ablation}~shows an ablation study of ~\modelnameshort~on the development set of ShARC. 

\paragraph{Retrieval outperforms word generation.}
BERTQA (``-edit, entail, extract''), which~\modelnameshort~reduces to after removing the editor, entailment, and extraction modules, presents a strong baseline that exceeds previous results on all metrics except for BLEU1.
This variant inquires about spans extracted from the text, which, while more relevant as indicated by the higher BLEU4 score, does not have the natural qualities of a question, hence it has a lower BLEU1.
Nonetheless, the large gains of BERTQA over the attentive Seq2Seq model shows that retrieval is a more promising technique for asking follow-up questions than word-by-word generation.
Similar findings were reported for question answering by~\citet{yatskar2019Qualitative}.

\paragraph{Extraction of document structure facilitates generalization.}
Adding explicit extraction of rules in the document (``-edit, entail'') forces the model to interpret all rules in the document versus only focusing on extracting the next inquiry.
This results in better performance in both decision classification and inquiry relevance compared to the variant that is not forced to interpret all rules.

\paragraph{Modeling entailment improves rule retrieval.}
The ``-edit'' model explicitly models whether an extracted rule is entailed by the user scenario and previous turns.
Modeling entailment allows the model to better predict whether a rule is entailed, and thus more often inquire about rules that are not entailed.
Figure~\ref{fig:pred1} illustrates one such example in which both extracted rules have high entailment score, and the model chooses to conclude the dialogue by answering \texttt{no} instead of making further inquiries.
Adding entailment especially improves in BLEU4 score, as the inquiries made by the model are more relevant and appropriate.

\paragraph{Editing retrieved rules results in more fluid questions.}
While~\modelnameshort~without the editor is able to retrieve rules that are relevant, these spans are not fluent questions that can be presented to the user.
The editor is able to edit the extracted rules into more fluid and coherent questions, which results further gains particularly in BLEU1.

\subsection{Error analysis}
In addition to ablation studies, we analyze errors~\modelnameshort~makes on the development set of ShARC.

\begin{figure}[!t]
    \centering
    \includegraphics[width=0.90\linewidth]{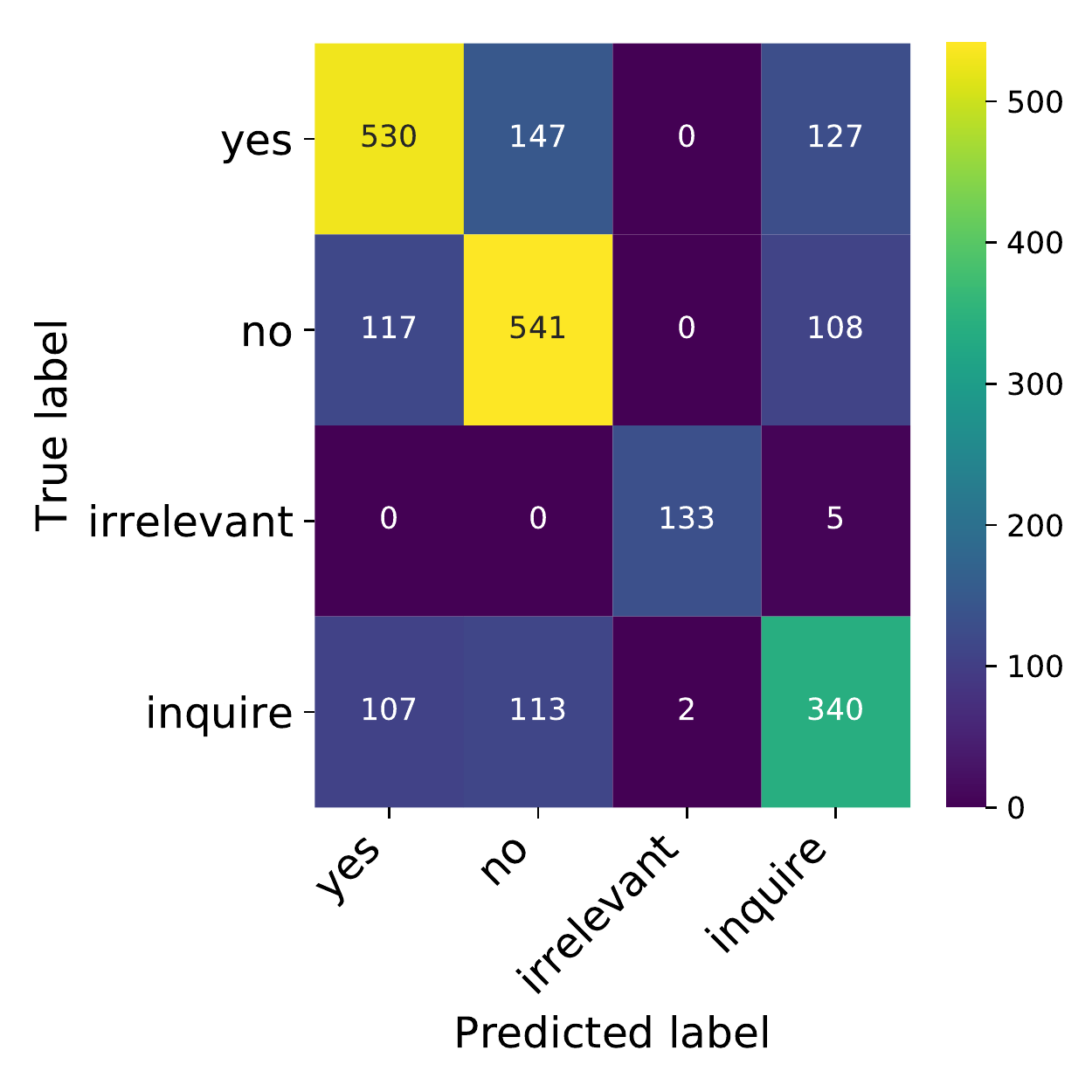}
    \vspace{-0.2in}
    \caption{
    Confusion matrix of decision predictions on the development set of ShARC.
    }
    \label{fig:confusion}
    \vspace{-0.2in}
\end{figure}

\paragraph{Decision errors.}
Figure~\ref{fig:confusion} shows the confusion matrix of decisions.
We specifically examine examples in which~\modelnameshort~produces an incorrect decision.
On the ShARC development set there are 726 such cases, which correspond to a 32.0\% error rate.
We manually analyze 100 such examples to identify commons types of errors.
Within these, in 23\% of examples, the model attempts to answer the user's initial question without resolving a necessary rule despite successfully extracting the rule.
In 19\% of examples, the model identifies and inquires about all necessary rules but comes to the wrong conclusion.
In 18\% of examples, the model makes a redundant inquiry about a rule that is entailed.
In 17\% of examples, the rule text contains ambiguous rules.
Figure~\ref{fig:pred3} contains one such example in which the annotator identified the rule ``a female Vietnam Veteran'', while the model extracted an alternative longer rule ``a female Vietnam Veteran with a child who has a birth defect''.
Finally, in 13\% of examples, the model fails to extract some rule from the document.
Other less common forms of errors include failures by the entailment module to perform numerical comparison, complex rule procedures that are difficult to deduce, and implications that require world knowledge.
These results suggests that improving the decision process after rule extraction is an important area for future work.

\paragraph{Inquiry quality.}
On 340 examples (15\%) in the ShARC development set,~\modelnameshort~generates an inquiry when it is supposed to.
We manually analyze 100 such examples to gauge the quality of generated inquiries.
On 63\% of examples, the model generates an inquiry that matches the ground-truth.
On 14\% of examples, the model makes inquires in a different order than the annotator.
On 12\% of examples, the inquiry refers to an incorrect subject (e.g.~``are you born early'' vs. ``is your baby born early''.
This usually results from editing an entity-less bullet point (``* born early'').
On 6\% of examples, the inquiry is lexically similar to the ground truth but has incorrect semantics (e.g.~``do you need savings'' vs. ``is this information about your savings'').
Again, this tends to result from editing short bullet points (e.g.~``* savings'').
These results indicate that when the model correctly chooses to inquire, it largely inquires about the correct rule.
They also highlight a difficulty in evaluating CMR --- there can be several correct orderings of inquiries for a document.

\section{Conclusion}

We proposed the~\modelname~(\modelnameshort), a conversational machine reading model that extracts implicit decision rules from text, computes whether each rule is entailed by the conversation history, inquires about rules that are not entailed, and answers the user's question.
\modelnameshort~achieved a new~\sota~result on the ShARC CMR dataset, outperforming existing systems as well as a new extractive QA baseline based on BERT.
In addition to achieving strong performance, we showed that~\modelnameshort~provides a more explainable alternative to prior work which do not model document structure.

\section*{Acknowledgments}
This research was supported in part by the ARO (W911NF-16-1-0121) and the NSF (IIS-1252835, IIS-1562364).
We thank Terra Blevins, Sewon Min, and our anonymous reviewers for helpful feedback.

\bibliography{acl2019}

\begin{thebibliography}{33}
\expandafter\ifx\csname natexlab\endcsname\relax\def\natexlab#1{#1}\fi

\bibitem[{Angeli et~al.(2015)Angeli, Johnson~Premkumar, and
  Manning}]{angeli2015openie}
Gabor Angeli, Melvin~Jose Johnson~Premkumar, and Christopher~D. Manning. 2015.
\newblock Leveraging linguistic structure for open domain information
  extraction.
\newblock In \emph{ACL}.

\bibitem[{Angeli and Manning(2014)}]{angeli2014Naturalli}
Gabor Angeli and Christopher~D. Manning. 2014.
\newblock Naturalli: Natural logic inference for common sense reasoning.
\newblock In \emph{EMNLP}.

\bibitem[{Bahdanau et~al.(2015)Bahdanau, Cho, and Bengio}]{bahdanau2014neural}
Dzmitry Bahdanau, Kyunghyun Cho, and Yoshua Bengio. 2015.
\newblock Neural machine translation by jointly learning to align and
  translate.
\newblock In \emph{ICLR}.

\bibitem[{Bordes et~al.(2013)Bordes, Usunier, Garcia-Duran, Weston, and
  Yakhnenko}]{bordes2013TransE}
Antoine Bordes, Nicolas Usunier, Alberto Garcia-Duran, Jason Weston, and Oksana
  Yakhnenko. 2013.
\newblock Translating embeddings for modeling multi-relational data.
\newblock In \emph{NIPS}.

\bibitem[{Choi et~al.(2018)Choi, He, Iyyer, Yatskar, Yih, Choi, Liang, and
  Zettlemoyer}]{choi2018QuAC}
Eunsol Choi, He~He, Mohit Iyyer, Mark Yatskar, Wen-tau Yih, Yejin Choi, Percy
  Liang, and Luke Zettlemoyer. 2018.
\newblock {QuAC}: Question answering in context.
\newblock In \emph{EMNLP}.

\bibitem[{Dettmers et~al.(2018)Dettmers, Pasquale, Pontus, and
  Riedel}]{dettmers2018conve}
Tim Dettmers, Minervini Pasquale, Stenetorp Pontus, and Sebastian Riedel. 2018.
\newblock Convolutional {2D} knowledge graph embeddings.
\newblock In \emph{AAAI}.

\bibitem[{Devlin et~al.(2019)Devlin, Chang, Lee, and
  Toutanova}]{Devlin2018BERT}
Jacob Devlin, Ming-Wei Chang, Kenton Lee, and Kristina Toutanova. 2019.
\newblock {BERT}: Pre-training of deep bidirectional transformers for language
  understanding.
\newblock In \emph{NAACL}.

\bibitem[{Henderson et~al.(2014)Henderson, Thomson, and Williams}]{dstc2}
Matthew Henderson, Blaise Thomson, and Jason~D Williams. 2014.
\newblock The second dialog state tracking challenge.
\newblock In \emph{SIGDIAL}.

\bibitem[{Hochreiter and Schmidhuber(1997)}]{Hochreiter1997LongSM}
Sepp Hochreiter and J{\"u}rgen Schmidhuber. 1997.
\newblock Long short-term memory.
\newblock \emph{Neural Computation}.

\bibitem[{Kingma and Ba(2015)}]{Kingma2014AdamAM}
Diederik~P. Kingma and Jimmy Ba. 2015.
\newblock Adam: A method for stochastic optimization.
\newblock In \emph{ICLR}.

\bibitem[{Lei et~al.(2016)Lei, Barzilay, and Jaakkola}]{lei2016rationalizing}
Tao Lei, Regina Barzilay, and Tommi Jaakkola. 2016.
\newblock Rationalizing neural predictions.
\newblock In \emph{EMNLP}.

\bibitem[{Lin et~al.(2018)Lin, Socher, and Xiong}]{lin2018multihopKG}
Xi~Victoria Lin, Richard Socher, and Caiming Xiong. 2018.
\newblock Multi-hop knowledge graph reasoning with reward shaping.
\newblock In \emph{EMNLP}.

\bibitem[{Manning et~al.(2014)Manning, Surdeanu, Bauer, Finkel, Bethard, and
  McClosky}]{Manning2014TheSC}
Christopher~D. Manning, Mihai Surdeanu, John Bauer, Jenny~Rose Finkel, Steven
  Bethard, and David McClosky. 2014.
\newblock The {Stanford CoreNLP} natural language processing toolkit.
\newblock In \emph{ACL}.

\bibitem[{Mintz et~al.(2009)Mintz, Bills, Snow, and
  Jurafsky}]{Mintz2009DistantSupRE}
Mike Mintz, Steven Bills, Rion Snow, and Daniel Jurafsky. 2009.
\newblock Distant supervision for relation extraction without labeled data.
\newblock In \emph{ACL}.

\bibitem[{{Moulin} and {Rousseau}(1992)}]{Moulin1992Automated}
B.~{Moulin} and D.~{Rousseau}. 1992.
\newblock Automated knowledge acquisition from regulatory texts.
\newblock \emph{IEEE Expert}.

\bibitem[{Mrk{\v{s}}i{\'c} et~al.(2017)Mrk{\v{s}}i{\'c}, S{\'e}aghdha, Wen,
  Thomson, and Young}]{mrkvsic2016neural}
Nikola Mrk{\v{s}}i{\'c}, Diarmuid~O S{\'e}aghdha, Tsung-Hsien Wen, Blaise
  Thomson, and Steve Young. 2017.
\newblock Neural belief tracker: Data-driven dialogue state tracking.
\newblock In \emph{ACL}.

\bibitem[{Parikh et~al.(2016)Parikh, T{\"a}ckstr{\"o}m, Das, and
  Uszkoreit}]{parikh2016dam}
Ankur Parikh, Oscar T{\"a}ckstr{\"o}m, Dipanjan Das, and Jakob Uszkoreit. 2016.
\newblock A decomposable attention model for natural language inference.
\newblock In \emph{EMNLP}.

\bibitem[{Pennington et~al.(2014)Pennington, Socher, and
  Manning}]{Pennington2014GloveGV}
Jeffrey Pennington, Richard Socher, and Christopher~D. Manning. 2014.
\newblock {GloVe}: Global vectors for word representation.
\newblock In \emph{EMNLP}.

\bibitem[{Press and Wolf(2017)}]{press2017WeightTying}
Ofir Press and Lior Wolf. 2017.
\newblock Using the output embedding to improve language models.
\newblock In \emph{ACL}.

\bibitem[{Rajpurkar et~al.(2016)Rajpurkar, Zhang, Lopyrev, and
  Liang}]{Rajpurkar2016SQuAD10}
Pranav Rajpurkar, Jian Zhang, Konstantin Lopyrev, and Percy Liang. 2016.
\newblock {SQuAD}: 100, 000+ questions for machine comprehension of text.
\newblock In \emph{EMNLP}.

\bibitem[{Reddy et~al.(2019)Reddy, Chen, and Manning}]{reddy2019coqa}
Siva Reddy, Danqi Chen, and Christopher~D Manning. 2019.
\newblock {CoQA}: A conversational question answering challenge.
\newblock \emph{TACL}.

\bibitem[{Riedel et~al.(2013)Riedel, Yao, McCallum, and
  Marlin}]{riedel2013REMatFac}
Sebastian Riedel, Limin Yao, Andrew McCallum, and Benjamin~M. Marlin. 2013.
\newblock Relation extraction with matrix factorization and universal schemas.
\newblock In \emph{NAACL}.

\bibitem[{Saeidi et~al.(2018)Saeidi, Bartolo, Lewis, Singh, Rockt{\"a}schel,
  Sheldon, Bouchard, and Riedel}]{Saeidi2018interpretation}
Marzieh Saeidi, Max Bartolo, Patrick Lewis, Sameer Singh, Tim Rockt{\"a}schel,
  Mike Sheldon, Guillaume Bouchard, and Sebastian Riedel. 2018.
\newblock Interpretation of natural language rules in conversational machine
  reading.
\newblock In \emph{EMNLP}.

\bibitem[{Srivastava et~al.(2014)Srivastava, Hinton, Krizhevsky, Sutskever, and
  Salakhutdinov}]{Srivastava14dropout}
Nitish Srivastava, Geoffrey Hinton, Alex Krizhevsky, Ilya Sutskever, and Ruslan
  Salakhutdinov. 2014.
\newblock Dropout: A simple way to prevent neural networks from overfitting.
\newblock \emph{JMLR}.

\bibitem[{Su et~al.(2016)Su, Gasic, Mrk{\v{s}}i{\'{c}}, Rojas~Barahona, Ultes,
  Vandyke, Wen, and Young}]{Su2016ActiveReward}
Pei-Hao Su, Milica Gasic, Nikola Mrk{\v{s}}i{\'{c}}, Lina~M. Rojas~Barahona,
  Stefan Ultes, David Vandyke, Tsung-Hsien Wen, and Steve Young. 2016.
\newblock On-line active reward learning for policy optimisation in spoken
  dialogue systems.
\newblock In \emph{ACL}.

\bibitem[{Vaswani et~al.(2017)Vaswani, Shazeer, Parmar, Uszkoreit, Jones,
  Gomez, Kaiser, and Polosukhin}]{vaswani2017Attention}
Ashish Vaswani, Noam Shazeer, Niki Parmar, Jakob Uszkoreit, Llion Jones,
  Aidan~N. Gomez, Lukasz Kaiser, and Illia Polosukhin. 2017.
\newblock Attention is all you need.
\newblock In \emph{NIPS}.

\bibitem[{Wen et~al.(2015)Wen, Gasic, Mrk{\v{s}}i{\'{c}}, Su, Vandyke, and
  Young}]{Wen2015SemanticConditionedGen}
Tsung-Hsien Wen, Milica Gasic, Nikola Mrk{\v{s}}i{\'{c}}, Pei-Hao Su, David
  Vandyke, and Steve Young. 2015.
\newblock Semantically conditioned lstm-based natural language generation for
  spoken dialogue systems.
\newblock In \emph{EMNLP}.

\bibitem[{Wen et~al.(2017)Wen, Vandyke, Mrk\v{s}i\'{c}, Ga{\v{s}}i{\'c},
  Rojas~Barahona, Su, Ultes, and
  Young}]{wen2017NetworkBasedEndToEndDialogueSystem}
Tsung-Hsien Wen, David Vandyke, Nikola Mrk\v{s}i\'{c}, Milica Ga{\v{s}}i{\'c},
  Lina~M. Rojas~Barahona, Pei-Hao Su, Stefan Ultes, and Steve Young. 2017.
\newblock A network-based end-to-end trainable task-oriented dialogue system.
\newblock In \emph{EACL}.

\bibitem[{Williams et~al.(2013)Williams, Raux, Ramachandran, and Black}]{dstc1}
Jason~D Williams, Antoine Raux, Deepak Ramachandran, and Alan Black. 2013.
\newblock The dialog state tracking challenge.
\newblock In \emph{SIGDIAL}.

\bibitem[{Wu et~al.(2016)Wu, Schuster, Chen, Le, Norouzi, Macherey, Krikun,
  Cao, Gao, Macherey, Klingner, Shah, Johnson, Liu, Kaiser, Gouws, Kato, Kudo,
  Kazawa, Stevens, Kurian, Patil, Wang, Young, Smith, Riesa, Rudnick, Vinyals,
  Corrado, Hughes, and Dean}]{Wu2016GooglesNM}
Yonghui Wu, Mike Schuster, Zhifeng Chen, Quoc~V. Le, Mohammad Norouzi, Wolfgang
  Macherey, Maxim Krikun, Yuan Cao, Qin Gao, Klaus Macherey, Jeff Klingner,
  Apurva Shah, Melvin Johnson, Xiaobing Liu, Lukasz Kaiser, Stephan Gouws,
  Yoshikiyo Kato, Taku Kudo, Hideto Kazawa, Keith Stevens, George Kurian,
  Nishant Patil, Wei Wang, Cliff Young, Jason Smith, Jason Riesa, Alex Rudnick,
  Oriol Vinyals, Gregory~S. Corrado, Macduff Hughes, and Jeffrey Dean. 2016.
\newblock Google's neural machine translation system: Bridging the gap between
  human and machine translation.
\newblock \emph{CoRR}, abs/1609.08144.

\bibitem[{Yatskar(2019)}]{yatskar2019Qualitative}
Mark Yatskar. 2019.
\newblock A qualitative comparison of coqa, squad 2.0 and quac.
\newblock In \emph{NAACL}.

\bibitem[{Young et~al.(2013)Young, Ga{\v{s}}i{\'c}, Thomson, and
  Williams}]{young2013POMDPDialogueReview}
Steve Young, Milica Ga{\v{s}}i{\'c}, Blaise Thomson, and Jason~D Williams.
  2013.
\newblock {POMDP}-based statistical spoken dialog systems: A review.
\newblock \emph{Proceedings of the IEEE}.

\bibitem[{Zhong et~al.(2018)Zhong, Xiong, and Socher}]{zhong2018global}
Victor Zhong, Caiming Xiong, and Richard Socher. 2018.
\newblock Global-locally self-attentive dialogue state tracker.
\newblock In \emph{ACL}.

\end{thebibliography}
\bibliographystyle{acl_natbib}

\clearpage

\appendix

\section{Appendices}
\subsection{BertQA Baseline}
\label{sec:bertqa}

Our BertQA baseline follows that proposed by~\citet{Devlin2018BERT} for the Stanford Question Answering Dataset (SQuAD)~\citep{Rajpurkar2016SQuAD10}.
Due to the differences in context between ShARC and SQuAD, we augment the input to the BERTQA model in a manner similar to Section~\ref{sec:extraction}.
The distinction here is that we additionally add the decision types ``yes'', ``no'', and ``irrelevant'' as parts of the input such that the problem is fully solvable via span extraction.
Similar to Section~\ref{sec:extraction}, let $U$ denote the BERT encoding of the length-$n$ input sequence.
The BERTQA model predicts a start score $s$ and an end score $e$. 

\begin{eqnarray}
s &= \mathrm{softmax}(U W_s + b_s) \in \real{n}\\
e &= \mathrm{softmax}(U W_e + b_e) \in \real{n}
\end{eqnarray}

We take the answer as the span $(i, j)$ that gives the highest score $s_i e_j$ such that $j >= i$.
Because we augment the input with decision labels, the model can be fully supervised via extraction endpoints.

\subsection{Creating noisy supervision for span extraction via span matching}
\label{sec:matching}

The ShARC dataset is constructed from full dialogue trees in which annotators exhaustively annotate yes/no branches of follow-up questions.
Consequently, each rule required to answer the initial user question forms a follow-up question in the full dialogue tree.
In order to identify rule spans in the document, we first reconstruct the dialogue trees for all training examples in ShARC.
For each document, we trim each follow-up question in its corresponding dialogue tree by removing punctuation and stop words.
For each trimmed question, we find the shortest best-match span in the document that has the least edit distance from the trimmed question, which we take as the corresponding rule span.
In addition, we extract similarly trimmed bullet points from the document as rule spans.
Finally, we deduplicate the rule spans by removing those that are fully covered by a longer rule span.
Our resulting set of rule spans are used as noisy supervision for the rule extraction module.
This preprocessing code is included with our code release.

\end{document}